\documentclass[runningheads]{llncs}
\usepackage{graphicx}
\usepackage[colorlinks=true,linkcolor=blue]{hyperref}

\usepackage{amsfonts}
\usepackage{amsmath}

\usepackage{array}
\usepackage{multirow}

\begin{document}
\title{MIM-OOD: Generative Masked Image Modelling for Out-of-Distribution Detection in Medical Images}
\titlerunning{MIM-OOD: Generative MIM for OOD in Medical Images}
%
\author{Sergio {Naval Marimont}\inst{1}\orcidID{0000-0002-7075-5586} \and Vasilis Siomos \inst{1} \orcidID{0009-0003-0985-2672} \and 
Giacomo Tarroni\inst{1, 2}\orcidID{0000-0002-0341-6138}}
%


\authorrunning{S. Naval Marimont et al.}
%
\institute{CitAI Research Centre, City, University of London, London, UK \and
BioMedIA, Imperial College, London, UK
\email{\{sergio.naval-marimont,vasilis.siomos,giacomo.tarroni\}@city.ac.uk}}
\maketitle              
\begin{abstract}
Unsupervised Out-of-Distribution (OOD) detection consists in identifying anomalous regions in images leveraging only models trained on images of healthy anatomy. An established approach is to \textit{tokenize} images and model the distribution of tokens with Auto-Regressive (AR) models. AR models are used to 1) identify anomalous tokens and 2) in-paint anomalous representations with in-distribution tokens. However, AR models are slow at inference time and prone to \textit{error accumulation} issues which negatively affect OOD detection performance. Our novel method, MIM-OOD, overcomes both speed and error accumulation issues by replacing the AR model with two task-specific networks: 1) a transformer optimized to identify anomalous tokens and 2) a transformer optimized to in-paint anomalous tokens using masked image modelling (MIM). Our experiments with brain MRI anomalies show that MIM-OOD substantially outperforms AR models (DICE 0.458 vs 0.301) while achieving a nearly 25x speedup (9.5s vs 244s).

\end{abstract}
\section{Introduction}
Supervised deep learning approaches achieve state of the art performance in many medical image analysis tasks \cite{Litjens2017}, but they require large amounts of manual annotations by medical experts. The manual annotation process is expensive and time-consuming, can lead to errors and suffers from inter-operator variability. Furthermore, supervised methods are specific to the anomalies annotated, which is in contrast with the diversity of naturally occurring anomalies. These limitations severely hinder applications of deep learning methods in the clinical practice. 

Unsupervised Out-of-Distribution (OOD) detection methods propose to bypass these limitations by seeking to identify anomalies relying only on anomaly-free data. Generically, the so-called \textit{normative} distribution of healthy anatomy is modelled by leveraging a training dataset of healthy images, and anomalous regions in test images are identified if they differ from the learnt distribution. A common strategy consists in modelling the \textit{normative} distribution using generative models \cite{Baur2020}. In particular, recent two-stage approaches have shown promising results \cite{Wang2020,Naval2020,PinayaW2022a,PinayaW2022d}. In order to segment anomalies, a first stage encodes the image into discrete latent representations referred as \textit{tokens}, from a VQ-VAE \cite{Oord2017}. The second stage aims to model the likelihood of individual \textit{tokens}, so a low likelihood can be used to identify those tokens that are not expected in the distribution of normal anatomies. Auto-Regressive (AR) modelling is the most common approach to model latent representation distributions \cite{Oord2017}. Furthermore, once the anomalous \textit{tokens} are identified, the generative capabilities of the AR model allow to in-paint anomalous regions with in-distribution \textit{tokens}. By decoding the now in-distribution \textit{tokens}, it is possible to obtain \textit{healed} images, and anomalies can be localized with the pixel-wise residuals between original and \textit{healed} images \cite{Wang2020,Naval2020,PinayaW2022a,PinayaW2022d}. 

Although this strategy is effective, AR modelling requires defining an artificial order for the \textit{tokens} so the latent distribution can be modelled as a sequence, and consequently suffer from two important drawbacks: 1) inference/image generation requires iterating through the latent variable sequence, which is computationally expensive, and 2) the fixed sequence order leads to error accumulation issues \cite{PinayaW2022d}. Error accumulation issues occur when AR models find OOD \textit{tokens} early on the sequence and the \textit{healed}/sampled sequence diverges from the original image, causing normal \textit{tokens} to also be replaced.

\subsubsection{Contributions:} we propose MIM-OOD, a novel approach for OOD detection with generative models that overcomes the aforementioned issues to outperform equivalent AR models both in accuracy and speed. Our main contributions are:
\begin{itemize}
    \item Instead of a single model for both tasks, MIM-OOD consists of two bi-directional Transformer networks: 1) the \textbf{Anomalous Token Detector}, trained to identify anomalous \textit{tokens}, and 2) the \textbf{Generative Masked Visual Token Model (MVTM)}, trained using the masked image modelling (MIM) strategy and used to in-paint anomalous regions with in-distribution tokens. To our knowledge, it is the first time MIM is leveraged for this task. 
    \item We evaluated MIM-OOD on brain MRIs, where it substantially outperformed AR models in detecting gliomas anomalies from BRATS dataset (DICE 0.458 vs 0.301) while also requiring a fraction of the inference time (9.5s vs 244s).
\end{itemize}

\section{Related Works} 

Generative models have been at the core of the literature in unsupervised OOD detection in medical image analysis. Vanilla approaches using Variational Auto-Encoders (VAE) \cite{Kingma2013} are based on the assumption that VAEs trained on healthy data would not be able to reconstruct anomalous regions and consequently voxel-wise residuals could be used as Anomaly Score (AS) \cite{Baur2020}. In \cite{Zimmerer2019}, authors found that the KL-divergence term in the VAE loss function could be used to both detect anomalous samples and localise anomalies in pixel space using gradient ascent. Chen et al. \cite{Chen2020} suggested using the gradients of the VAE loss function w.r.t. pixels values to iteratively \textit{heal} the images and turn them into in-distribution samples, in a so-called \textit{restoration} process.  Approaches using Generative Adversarial Networks (GANs) \cite{Goodfellow2014} assume that anomalous samples are not encoded in the normative distribution and that AS can be derived from pixel-wise residuals between test samples and reconstructions \cite{Schlegl2019}. As introduced previously, the most recent approaches based on generative models rely on two-stage image modelling \cite{Wang2020,Naval2020,PinayaW2022a}: the first stage is generally a \textit{tokenizer} \cite{Oord2017} that encodes images in discrete latent representations (referred as \textit{tokens}), followed by a second stage that learns the \textit{normative} distribution of \textit{tokens} leveraging AR models. To overcome the limitations of AR modelling, Pinaya et al. \cite{PinayaW2022d} propose replacing AR with a latent diffusion model \cite{RombachR2022}.

A novel and efficient strategy to model the latent distribution is generative masked image modelling (MIM), which has been used for image generation \cite{ChangH2022}. MIM consists in 1) dividing the variables of a multivariate joint probability distribution into masked and visible subsets, and in 2) modelling the probability of masked variables given visible ones. Masked Image Modelling strategy, when applied to latent visual \textit{tokens} is referred to as Masked Visual Token Modelling (MVTM). MVTM produces high quality images by iteratively masking and resampling the latent variables where the model is less confident. In \cite{Lezama2022}, authors improve the quality of the generated images by training an additional model, named \textit{Token Critic}, to identify which latent variables require resampling by the MVTM model. The Token Critic is trained to identify which \textit{tokens} are sampled from the model vs which tokens were in the original image. Token Critic and MVTM address the tasks of identifying inconsistent tokens and in-painting masks tokens, respectively, which are similar to the roles of AR models in the unsupervised OOD detection literature. We took inspiration from these efficient strategies to design our novel MIM-OOD method.

\section{Method}
\subsection{Vector Quantized Variational Auto-Encoders}

Vector Quantized Variational Auto-Encoders (VQ-VAEs) \cite{Oord2017} encode images $x \in \mathbb{R}^{H \times W}$  into representations ${z_q \in \textbf{K}^{H / f \times W /f } }$ where $\textbf{K}$ is a set of discrete representations, referred as \textit{tokens}, and $f$ is a downsampling factor measuring the spatial compression between pixels and \textit{tokens}. A VQ-VAE first encodes the images to a continuous space  ${z_e \in \mathbb{R}^{D \times H / f \times W /f }}$. Continuous representations are then discretized using an embedding space with $\lvert \textbf{K} \rvert$ embeddings $e_k \in \mathbb{R}^D$. Specifically, \textit{tokens} are defined as the index of the embedding vector $e_k$ nearest to $z_e$: $z_q = argmin_j \lVert z_e-e_j \rVert_2$. For a detailed explanation, please refer to the original paper \cite{Oord2017}.

\subsection{Generative Masked Visual Token Modelling (MVTM)} 
Masked modelling consists in learning the probability distribution of a set of occluded variables based on a set of observed ones from a given multivariate distribution. Occlusions are produced by replacing the values of variables with a special $\left[ MASK \right]$ \textit{token}. Consequently, the task is to learn $p(y_M \mid y_U)$, where $y_M, y_U$ are the masked and unmasked exclusive subsets of $Y$. The binary mask $\textbf{m} = \left[ m_i \right]^N_{i=1}$  defines which variables are masked: if $i \in M$ then $m_i = 1$ and $y_i$ is replaced with $\left[ MASK \right]$. The training objective is to maximize the marginal cross-entropy for each masked \textit{token}:

\begin{equation} \label{eq:2}
    \mathcal{L}_{MVTM} = \sum_{ \forall y_i \in Y_M} \log p_{\phi}(y_i \mid Y_U)
\end{equation} 

\begin{figure}[b!]
\includegraphics[width=\textwidth]{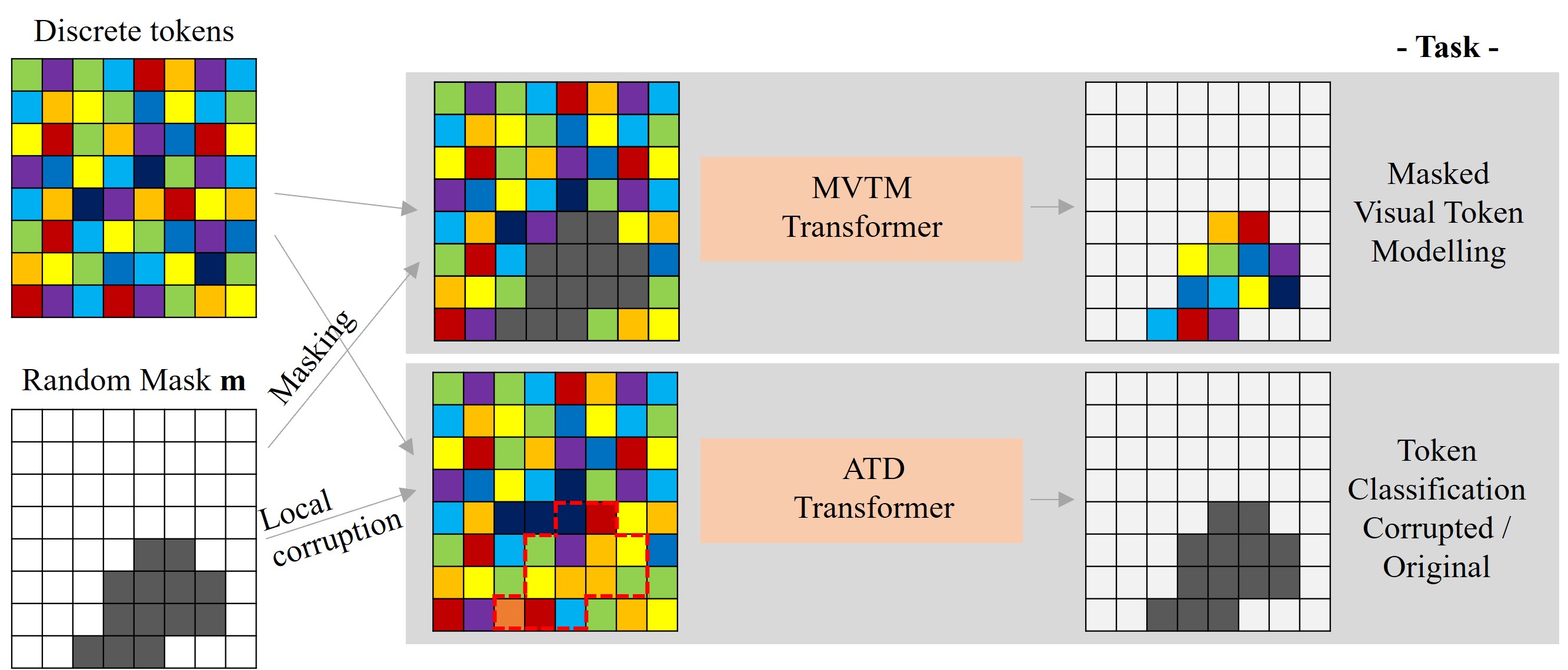}
\caption{Training tasks diagram. Given a random mask $\textbf{m}$ and an healthy image representation (generated by the VQ-VAE), two Transformers are trained. Upper: MVTM is trained to predict masked \textit{tokens}. Lower: Anomalous Token Detector (ATD) is trained to identify locally corrupted \textit{tokens}.} \label{fig1}
\end{figure}

In MVTM, $Y$ is the set of \textit{tokens} ${z_q \in \textbf{K}^{H / f \times W /f }}$ encoding image $x$. We use a multi-layer bi-directional Transformer to model the probability $p(y_i \mid Y_U)$ given the masked input. We optimize Transformer weights $\phi$ using back-propagation to minimize the cross-entropy between the ground-truth \textit{tokens} and predicted \textit{token} for the masked variables. The upper section in Figure \ref{fig1} describes the MVTM task. 

The MVTM model can be leveraged to both in-paint latent regions using its generative capabilities and to identify anomalous \textit{tokens}. A naive approach to identify anomalous \textit{tokens} would be to use the predicted $p(y_i \mid Y_U)$ to identify \textit{tokens} with low likelihood. However, by optimising over the marginals for each masked \textit{token}, the model learns the distribution of each of the masked variables independently and fails to model the \textit{joint} distribution of masked \textit{tokens} \cite{Lezama2022}. Given the above limitation we introduce a second latent model specialized to identify anomalous \textit{tokens}.

\subsection{Anomalous Token Detector (ATD)}
The Anomalous Token Detector (ATD) is trained to identify local corruptions, similar to the task introduced in \cite{Tan2020} but in latent space instead of pixel-space. We create the corruptions by replacing the \textit{tokens} in mask $\textbf{m}$ with random \textit{tokens}. Consequently, our ATD bi-directional Transformer $\theta$ receives as input $ \tilde y = y \odot (1 - m) + r \odot m  $, where $r$ is a random set of \textit{tokens} and $\odot$ is element-wise multiplication. It is trained to minimize a binary cross-entropy objective: 

\begin{equation} \label{eq:3}
    \mathcal{L}_{TC} = \sum_{i}^{N} m_i \log  p_{\theta}(y_i) + (1-m_i) \log (1 - p_{\theta}(y_i))
\end{equation} 

The bottom section in Figure \ref{fig1} describes the proposed ATD task. Both the MVTM and ATD masks were generated by a random walk of a brush with a randomly changing width. 

\subsection{Image restoration procedure}
At inference time, our goal is to evaluate if an image is consistent with the learnt healthy distribution and, if not, heal it by applying local transformations that replace anomalies with healthy tissue to generate a \textit{restoration}. We can then localise anomalies by comparing the restored and original images. The complete MIM-OOD pipeline consists of the following steps:
\begin{enumerate}
    \item \textbf{Tokenise} the image using the VQ-VAE Encoder.
    \item \textbf{Identify anomalous tokens} by selecting \textit{tokens} with an ATD prediction score greater than a threshold $\lambda$. Supplementary Materials include validation set results for different $\lambda$ values.
    \item \textbf{Restore anomalous tokens} by in-painting anomalous \textit{tokens} with generative masked modelling. We sample tokens based on the likelihood assigned by MVTM: $ \hat y_t \sim p_{\phi}(y_t \mid Y_U)$ and replace original tokens with samples in masked positions: $ y_{t+1} = y_{t} \odot (1 - m) + \hat y_t \odot m  $.
    \item \textbf{Decode} \textit{tokens} using the VQ-VAE Decoder to generate the restoration $x_T$.
    \item \textbf{Compute the Anomaly Score (AS)} as the pixel-wise residuals between restoration $x_T$ and original image $x_0$: $AS = \lvert x_T - x_0 \rvert $. AS is smoothed to remove edges with $min$ and $average-pooling$ filters\cite{Naval2020}. 
\end{enumerate}

Note that sampling from MVTM is done independently for each \textit{token}, so it is possible that sampled \textit{tokens} are inconsistent with each other. To address this issue, we can iterate $T$ times steps 2 and 3. Additionally, we can also generate $R$ multiple restorations per input image (i.e., repeating for each restoration step 3). We evaluated different values of $T$ steps and $R$ restorations, and validation set results are included in the Supplementary Materials. Figure \ref{fig2} describes the end-to-end inference pipeline.

\begin{figure}[t!]
\includegraphics[width=\textwidth]{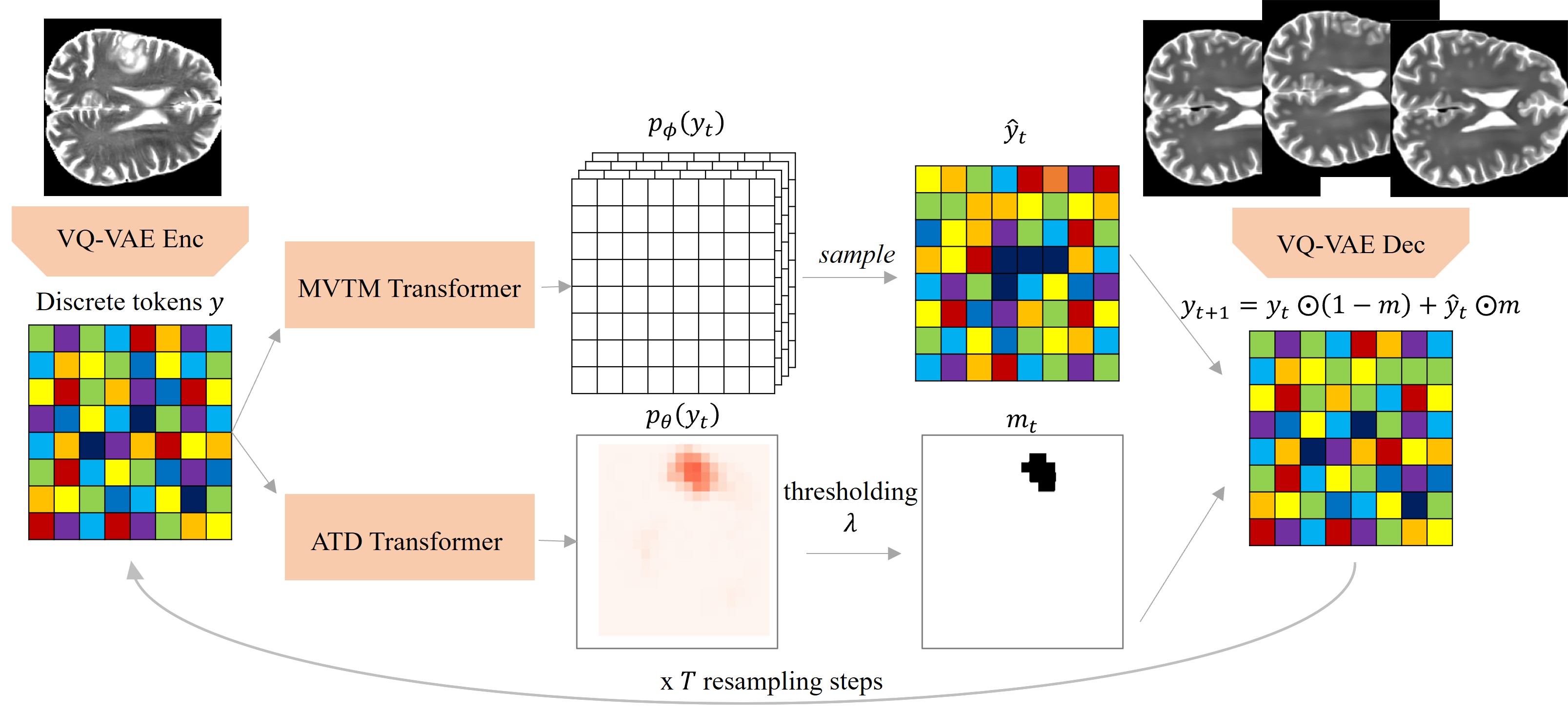}
\caption{MIM-OOD pipeline: the Anomalous Token Detector predicts the likelihood of \textit{tokens} being anomalous. \textit{Tokens} with ATD prediction score $> \lambda$ are deemed anomalous and replaced with samples from the MVTM model. We perform $T$ iterations of the above procedure in parallel to generate $R$ number of restorations.} \label{fig2}
\end{figure}

\section{Experiments and Results}
We evaluated MIM-OOD on brain MRIs (training on a dataset of normal images and testing on one with anomalies) and compared our approach to an AR benchmark thoroughly evaluated in the literature \cite{Naval2020,PinayaW2022a}. Two publicly available datasets were used:
\begin{itemize}
    \item The Human Connectome Project Young Adult (HCP) dataset \cite{VanEssen2012} with images of 1,113 young and healthy subjects which we split into a training set with 1013 images and a validation set with 100 images.
    \item The Multimodal Brain Tumor Image Segmentation Benchmark (BRATS) \cite{Menze2014}, from the 2021 challenge. The dataset contains images with gliomas and comes with ground-truth segmentation masks highlighting their location. We randomly selected 100 images as validation set and 200 images as test set.
\end{itemize}
From both datasets we obtained pre-processed, skull-stripped T2-weighted structural images. We re-sampled both datasets to a common isotropic spacing of 1 mm and obtained random axial slices of 160x160 pixels. Intensity values were clipped to percentile 98 and normalized to the range [0,1]. Latent models were only trained with random flip and rotations augmentations. We included broader-than-usual intensity augmentations during VQ-VAE training.

\subsubsection{Network architecture and implementation details:}
For the VQ-VAE we used the architecture from \cite{EsserP2021} and trained with MAE reconstruction loss for 300k steps. We used a codebook with $\lvert \textbf{K} \rvert =256$ and $D=256$ and a downsampling factor $f=8$. For the AR, ATD and MVTM we used a vanilla Transformer \cite{Vaswani2017} with depth 12, layer normalization, and a stochastic depth of $0.1$. Both the ATD and the MVTM Transformers are bi-directional since there is no sequence order to be enforced, in contrast to AR models. Transformers were trained for 200k steps.

We used AdamW with a cosine annealing scheduler, starting with learning rate of $5 \times 10^{-5}$ and weight decay $1 \times 10^{-5}$. Our MONAI \cite{monai} implementation and trained models are made publicly available in \footnote{\href{https://github.com/snavalm/MIM-OOD}{\texttt{https://github.com/snavalm/MIM-OOD}}}. 

\subsubsection{Performance evaluation:} 
The role of the latent models (AR and MVTM) is both to identify anomalous latent variables and to replace them with in-distribution values for restoration. To evaluate the identification task, we setup a proxy task of classifying as anomalous \textit{tokens} corresponding to image areas where anomalies are present. To this end, we downsample annotated ground-truth labels by the VQ-VAE scaling factor $f = 8$. Using \textit{token} likelihood as a Anomaly Score (AS) in latent space we computed the Average Precision (AP) and the Area Under the Receiver Operating Characteristic curve (AUROC). Table \ref{Table:tab1} summarizes the proxy task results on the validation set. Our proposed approach relying on the ATD substantially outperforms the competitors.

\begin{table}
\caption{Results for Anomalous Token identification in BRATS validation set.}\label{Table:tab1}
\begin{tabular}{p{7cm} >{\centering\arraybackslash}p{2cm} >{\centering\arraybackslash}p{2cm}}
\hline
Method &  AP & AUROC \\
\hline
AR \cite{Naval2020} & 0.054 & 0.773 \\
MVTM \cite{ChangH2022} & 0.084 & 0.827 \\
MVTM + Token Critic \cite{Lezama2022} & 0.041 & 0.701 \\
MIM-OOD (Ours) & \textbf{0.186} & \textbf{0.859} \\
\end{tabular}
\end{table}

We then evaluate MIM-OOD's capability to localise anomalies in image space on the test set. We report the best achievable $\left[ DICE \right]$ score following the conventions in recent literature \cite{Baur2020,PinayaW2022a}. Additionally, we report AP, AUROC and inference time per batch of 32 images (IT (s)) using a single Nvidia RTX3090.

\begin{table}
\caption{Results for Pixel-wise Anomaly Detection in BRATS test set.}\label{Table:tab2}
\begin{tabular}{p{6cm} >{\centering\arraybackslash}p{1.3cm} >{\centering\arraybackslash}p{1.3cm} >{\centering\arraybackslash}p{1.3cm} >{\centering\arraybackslash}p{1.3cm}}
\hline
Method &  [DICE] & AP & AUROC & IT (s) \\
\hline
AR restoration $R=4$ ${}^{(1)}$ \cite{Naval2020} & 0.301 & 0.191 & 0.891 & 244 \\
MVTM + Token Critic R=4 \cite{Lezama2022} & 0.201 & 0.131 & 0.797 & 5.4 \\
MIM-OOD R=4 (Ours) ${}^{(2)}$ & 0.458 & 0.399 & 0.926 & 9.5 \\
MIM-OOD R=8 (Ours) ${}^{(2)}$ & \textbf{0.461} & \textbf{0.404} & \textbf{0.928} & 19.9 \\
\hline
\multicolumn{5}{l}{1 - Implementation from \cite{Naval2020} with 4 restorations and $\lambda_{NLL} = 6$, using Transformer} \\
\multicolumn{5}{l}{instead of PixelSNAIL AR architecture.} \\
\multicolumn{5}{l}{2 - Our approach with 4 and 8 restorations respectively, $\lambda = 0.005$ and $T=8$.} \\
\end{tabular}
\end{table}

The results in Table \ref{Table:tab2} show that MIM-OOD improves the $[DICE]$ score by 15 points (0.301 vs 0.458) when using the same number of restorations ($R=4$), while at the same time reducing the inference time (244s vs 9.5s for a batch of 32 images). These results are consistent with the previous anomalous \textit{token} identification task where we showed that our ATD model was able to better identify anomalous \textit{tokens} requiring restorations. Qualitative results are included in Figure \ref{fig3} and in Supplementary Materials. 

It is worth noting that our AR baseline slightly underperforms compared to previous published results (DICE of 0.301 vs 0.328 in BRATS dataset \cite{PinayaW2022a,PinayaW2022d}) due to different experimental setups: different modalities (T2 vs FLAIR), training sets (HCP with N=1,013 vs UK Biobank with N=14,000) and pre-processing pipelines. While both T2 and FLAIR modalities are common in the OOD literature, T2 was chosen for our experiments because of the availability of freely accessible anomaly-free datasets. The differences in experimental setups makes results not directly comparable, however our approach shows promising performance and high efficiency when comparing with both AR Ensemble (\cite{PinayaW2022a} DICE 0.537 and 4,907 seconds per 100 image batch) and Latent Diffusion (\cite{PinayaW2022d} DICE 0.469 and 324 seconds per 100 image batch). 

\begin{figure}
\centering
\includegraphics[width=0.9\textwidth]{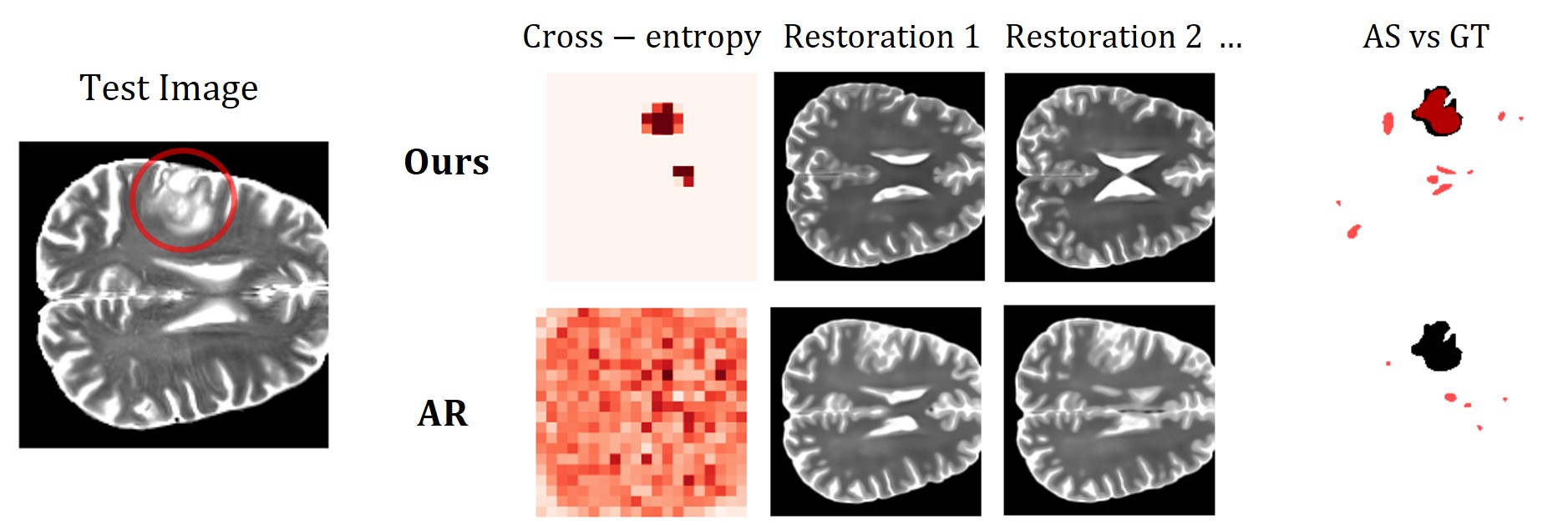}
\caption{Qualitative comparison. Our method outperforms the AR at both identifying the anomalous latent \textit{tokens} and generating restorations where the lesion has been healed. This translates to a more reliable AS map (pink) vs ground truth (black).} \label{fig3}
\end{figure}

\section{Conclusion}

We developed MIM-OOD, a novel unsupervised anomaly detection technique that, to our knowledge, leverages for the first time the concept of Generative Masked Modelling for this task. In addition, we introduce ATD, a novel approach to identify anomalous latent variables which synergizes with the MVTM to generate healed restorations. Our results show that our technique outperforms previous AR-based approaches in unsupervised glioma segmentation in brain MRI. In the future, we will perform further evaluations of our approach, testing it on data with other brain pathologies and using additional image modalities. 

%
%

%
%
%
\bibliographystyle{splncs04}
\bibliography{ref.bib}

\begin{thebibliography}{10}
\providecommand{\url}[1]{\texttt{#1}}
\providecommand{\urlprefix}{URL }
\providecommand{\doi}[1]{https://doi.org/#1}

\bibitem{Baur2020}
Baur, C., Denner, S., Wiestler, B., Albarqouni, S., Navab, N.: Autoencoders for
  {Unsupervised} {Anomaly} {Segmentation} in {Brain} {MR} {Images}: {A}
  {Comparative} {Study}. Medical image analysis (02 Jan 2021),  69:101952
  (2021)

\bibitem{Menze2014}
Bjoern H~Menze, e.a.: The {Multimodal} {Brain} {Tumor} {Image} {Segmentation}
  {Benchmark} ({BRATS}). IEEE Trans Med Imaging. 2015 Oct;34(10):1993-2024.doi:
  10.1109/TMI.2014.2377694. Epub 2014 Dec 4.  (2014)

\bibitem{monai}
Cardoso, M.J., et. al.: {MONAI}: {An} open-source framework for deep learning
  in healthcare (Nov 2022), arXiv:2211.02701 [cs]

\bibitem{ChangH2022}
Chang, H., Zhang, H., Jiang, L., Liu, C., Freeman, W.T.: Maskgit: {Masked}
  generative image transformer. In: Proceedings of the {IEEE}/{CVF}
  {Conference} on {Computer} {Vision} and {Pattern} {Recognition}. pp.
  11315--11325 (2022)

\bibitem{Chen2020}
Chen, X., You, S., Tezcan, K.C., Konukoglu, E.: Unsupervised {Lesion}
  {Detection} via {Image} {Restoration} with a {Normative} {Prior}. Proceedings
  of The 2nd International Conference on Medical Imaging with Deep Learning
  \textbf{PMLR 102},  540--556 (2020)

\bibitem{EsserP2021}
Esser, P., Rombach, R., Ommer, B.: Taming transformers for high-resolution
  image synthesis. In: Proceedings of the {IEEE}/{CVF} conference on computer
  vision and pattern recognition. pp. 12873--12883 (2021)

\bibitem{Goodfellow2014}
Goodfellow, I.J., Pouget-Abadie, J., Mirza, M., Xu, B., Warde-Farley, D.,
  Ozair, S., Courville, A., Bengio, Y.: Generative {Adversarial} {Networks}.
  Advances in neural information processing systems pp. 2672--2680 (2014)

\bibitem{Kingma2013}
Kingma, D.P., Welling, M.: Auto-{Encoding} {Variational} {Bayes}. The 2nd
  International Conference on Learning Representations (ICLR)  (2013)

\bibitem{Lezama2022}
Lezama, J., Chang, H., Jiang, L., Essa, I.: Improved masked image generation
  with token-critic. In: Computer {Vision}–{ECCV} 2022: 17th {European}
  {Conference}, {Tel} {Aviv}, {Israel}, {October} 23–27, 2022, {Proceedings},
  {Part} {XXIII}. pp. 70--86. Springer (2022)

\bibitem{Litjens2017}
Litjens, G., Kooi, T., Bejnordi, B.E., Setio, A.A.A., Ciompi, F., Ghafoorian,
  M., van~der Laak, J.A.W.M., van Ginneken, B., Sánchez, C.I.: A {Survey} on
  {Deep} {Learning} in {Medical} {Image} {Analysis}. Medical image analysis
  \textbf{vol. 42},  60--88 (2017)

\bibitem{Naval2020}
Naval~Marimont, S., Tarroni, G.: Anomaly detection through latent space
  restoration using vector quantized variational autoencoders. In: 2021 {IEEE}
  18th {International} {Symposium} on {Biomedical} {Imaging} ({ISBI}). pp.
  1764--1767. IEEE (2021)

\bibitem{Oord2017}
van~den Oord, A., Vinyals, O., Kavukcuoglu, K.: Neural {Discrete}
  {Representation} {Learning}. NIPS'17: Proceedings of the 31st International
  Conference on Neural Information Processing Systems pp. 6309--6318 (2017)

\bibitem{PinayaW2022d}
Pinaya, W.H.L., Graham, M.S., Gray, R., Da~Costa, P.F., Tudosiu, P.D., Wright,
  P., Mah, Y.H., MacKinnon, A.D., Teo, J.T., Jager, R., {others}: Fast
  {Unsupervised} {Brain} {Anomaly} {Detection} and {Segmentation} with
  {Diffusion} {Models}. arXiv preprint arXiv:2206.03461  (2022)

\bibitem{PinayaW2022a}
Pinaya, W.H.L., Tudosiu, P.D., Gray, R., Rees, G., Nachev, P., Ourselin, S.,
  Cardoso, M.J.: Unsupervised brain imaging {3D} anomaly detection and
  segmentation with transformers. Medical Image Analysis  \textbf{79},  102475
  (Jul 2022)

\bibitem{RombachR2022}
Rombach, R., Blattmann, A., Lorenz, D., Esser, P., Ommer, B.: High-resolution
  image synthesis with latent diffusion models. In: Proceedings of the
  {IEEE}/{CVF} {Conference} on {Computer} {Vision} and {Pattern} {Recognition}.
  pp. 10684--10695 (2022)

\bibitem{Schlegl2019}
Schlegl, T., Seeböck, P., Waldstein, S.M., Langs, G., Schmidt-Erfurth, U.:
  f-{AnoGAN}: {Fast} unsupervised anomaly detection with generative adversarial
  networks. Medical image analysis  \textbf{54},  30--44 (2019), publisher:
  Elsevier

\bibitem{Tan2020}
Tan, J., Hou, B., Batten, J., Qiu, H., Kainz, B.: Detecting {Outliers} with
  {Foreign} {Patch} {Interpolation}. Machine Learning for Biomedical Imaging
  \textbf{1}(April 2022 issue),  1--27 (Apr 2022)

\bibitem{VanEssen2012}
Van~Essen, D.C.e.a.: The {Human} {ConnectomeProject}: a data acquisition
  perspective. Neuroimage  \textbf{vol 62.4},  2222--2231. (2012)

\bibitem{Vaswani2017}
Vaswani, A., Shazeer, N., Parmar, N., Uszkoreit, J., Jones, L., Gomez, A.N.,
  Kaiser, {\textbackslash}., Polosukhin, I.: Attention is all you need.
  Advances in neural information processing systems  \textbf{30} (2017)

\bibitem{Wang2020}
Wang, L., Zhang, D., Guo, J., Han, Y.: Image {Anomaly} {Detection} {Using}
  {Normal} {Data} {Only} by {Latent} {Space} {Resampling}. Applied Sciences
  \textbf{10}(23), ~8660 (Jan 2020), number: 23 Publisher: Multidisciplinary
  Digital Publishing Institute

\bibitem{Zimmerer2019}
Zimmerer, D., Isensee, F., Petersen, J., Kohl, S., Maier-Hein, K.: Unsupervised
  {Anomaly} {Localization} using {Variational} {Auto}-{Encoders}. Medical Image
  Computing and Computer Assisted Intervention – MICCAI 2019. Lecture Notes
  in Computer Science  \textbf{vol 11767} (2019)

\end{thebibliography}

\newpage
\title{MIM-OOD: Generative Masked Image Modelling for Out-of-Distribution Detection in Medical Images}
\titlerunning{MIM-OOD: Generative MIM for OOD in Medical Images}
%
\author{Sergio {Naval Marimont}\inst{1}\orcidID{0000-0002-7075-5586} \and Vasilis Siomos \inst{1} \orcidID{0009-0003-0985-2672} \and 
Giacomo Tarroni\inst{1, 2}\orcidID{0000-0002-0341-6138}}
%


\authorrunning{S. Naval Marimont et al.}
%
\institute{CitAI Research Centre, City, University of London, London, UK \and
BioMedIA, Imperial College, London, UK
\email{\{sergio.naval-marimont,vasilis.siomos,giacomo.tarroni\}@city.ac.uk}}
\maketitle              

\section*{Supplementary Materials}

\begin{figure}[!h]
\includegraphics[width=\textwidth]{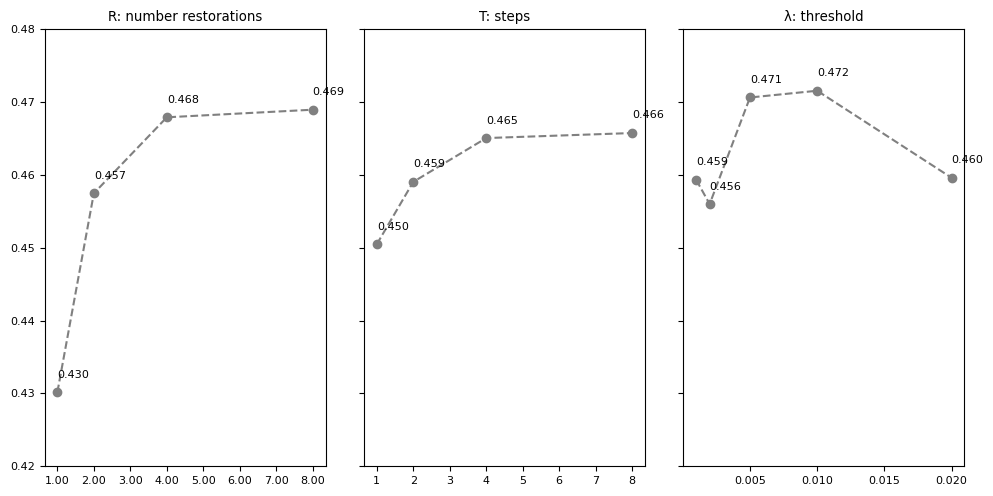}
\caption{Validation set $[DICE]$ results for hyperparameters evaluated.  Column 1: Number of restorations ($R$) with $T=4$ steps and $\lambda = 0.01$.   Column 2: Number of steps ($T$) with $R=4$ and $\lambda=0.01$. Column 3:  Threshold $\lambda$  with $R=4$ and $T=4$.} \label{figsm2}
\end{figure}

\begin{figure}[!h]
\includegraphics[width=\textwidth]{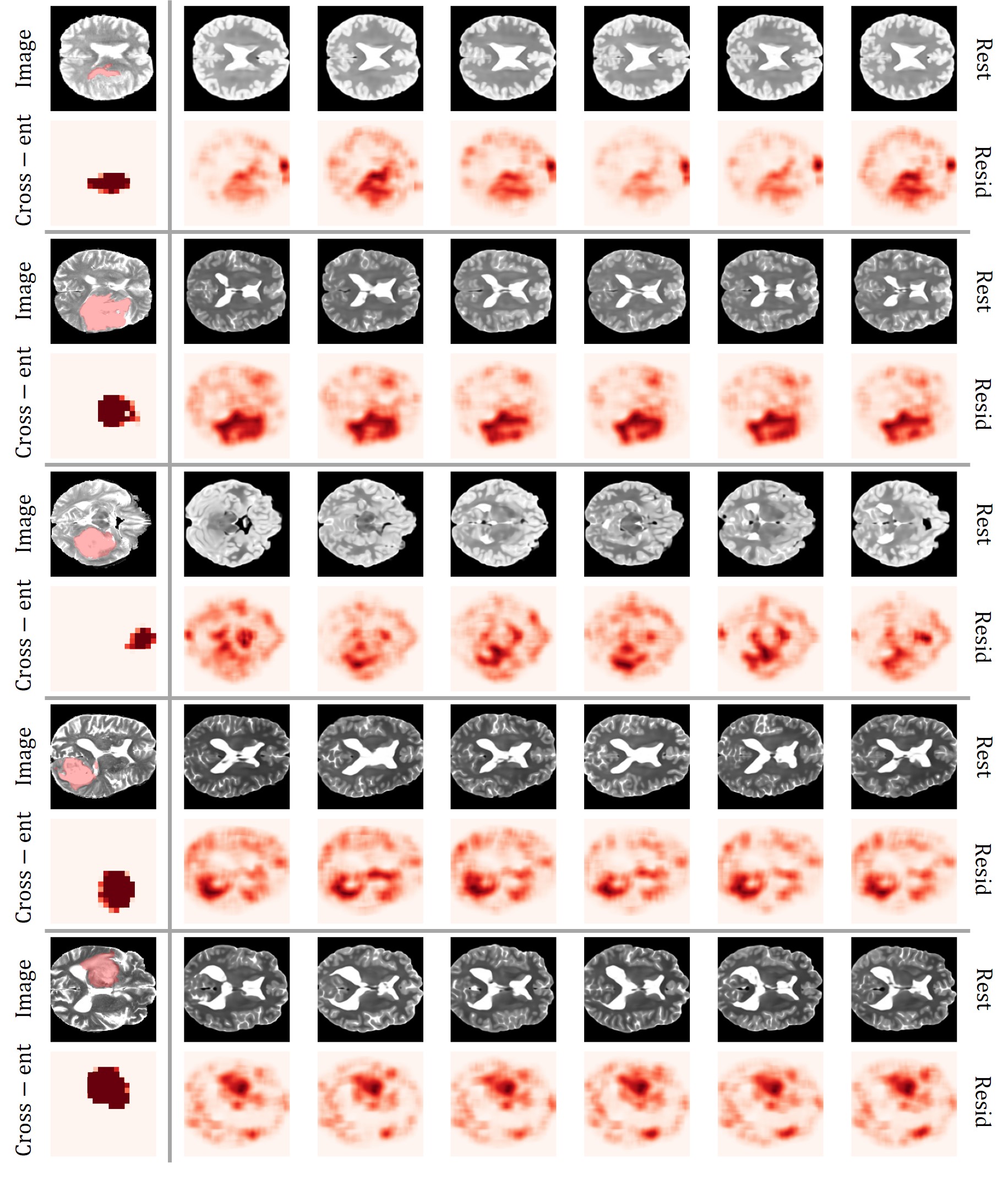}
\caption{Qualitative examples of restorations with the proposed method for 5 input images. First column shows test image (odd rows) and upsampled ATD output (even rows). Columns 2-7 show restorations (odd rows) and smoothed residuals between original image and restoration (even rows). Restorations obtained with $T=8$ steps and $\lambda=0.01$).} \label{figsm1}
\end{figure}

\end{document}